\documentclass[letterpaper]{article} 
\usepackage{aaai25}  
\usepackage{times}  
\usepackage{helvet}  
\usepackage{courier}  
\usepackage[hyphens]{url}  
\usepackage{graphicx} 
\usepackage{natbib}  
\usepackage{caption} 
\usepackage{algorithm}
\usepackage{algorithmic}
\usepackage{amsmath}
\usepackage{booktabs}
\usepackage{multirow}

\usepackage{newfloat}
\usepackage{listings}
\DeclareCaptionStyle{ruled}{labelfont=normalfont,labelsep=colon,strut=off} 
\floatstyle{ruled}
\newfloat{listing}{tb}{lst}{}
\floatname{listing}{Listing}
\title{Hierarchical Vector Quantization for Unsupervised Action Segmentation}
\author {
    Federico Spurio\textsuperscript{\rm 1,\rm 3},
    Emad Bahrami\textsuperscript{\rm 1,\rm 3},
    Gianpiero Francesca\textsuperscript{\rm 2},
    Juergen Gall\textsuperscript{\rm 1,\rm 3}
}
\affiliations {
    \textsuperscript{\rm 1}University of Bonn, Germany\\
    \textsuperscript{\rm 2}Toyota Motor Europe, Belgium\\
    \textsuperscript{\rm 3}Lamarr Institute for Machine Learning and Artificial Intelligence, Germany\\
    \{fspurio, bahrami, gall\}@iai.uni-bonn.de, gianpiero.francesca@toyota-europe.com
}

\newcommand{\ours}{HVQ}
\newcommand{\fvq}[0]{z_{j}}
\newcommand{\svq}[0]{q_{i}}

\newcommand{\bfm}[0]{54.4}
\newcommand{\bff}{39.7}
\newcommand{\bfr}{44.9}
\newcommand{\bfjs}{82.5}
\newcommand{\ytif}{35.1}
\newcommand{\ytim}{50.3}
\newcommand{\ytir}{38.7}
\newcommand{\ikeaf}{27.6}

\newcommand{\ikeatf}{30.7}
\newcommand{\ikeatm}{51.2}
\newcommand{\ikeajs}{64.8}

\DeclareMathOperator{\argmax}{argmax}

\newcommand{\figref}[1]{Fig.~\ref{#1}}
\newcommand{\tabref}[1]{Tab.~\ref{#1}}

\newcommand{\secref}[1]{Section~\ref{#1}}

\def\ie{\textit{i.e.}}
\def\eg{\textit{e.g.}}

\begin{document}

\maketitle

\begin{abstract}
In this work, we address unsupervised temporal action segmentation, which segments a set of long, untrimmed videos into semantically meaningful segments that are consistent across videos. While recent approaches combine representation learning and clustering in a single step for this task, they do not cope with large variations within temporal segments of the same class. To address this limitation, we propose a novel method, termed Hierarchical Vector Quantization (\ours), that consists of two subsequent vector quantization modules. This results in a hierarchical clustering where the additional subclusters cover the variations within a cluster. We demonstrate that our approach captures the distribution of segment lengths much better than the state of the art. To this end, we introduce a new metric based on the Jensen-Shannon Distance (JSD) for unsupervised temporal action segmentation.
We evaluate our approach on three public datasets, namely Breakfast, YouTube Instructional and IKEA ASM. Our approach outperforms the state of the art in terms of F1 score, recall and JSD.  
\end{abstract}

\begin{links}
    \link{Code}{https://github.com/FedeSpu/HVQ}
\end{links}

%

\section{Introduction}
\label{sec:intro}

Analyzing videos at a temporal level is crucial for applications in domains like healthcare, manufacturing, neuroscience, and robotics. However, fully-supervised methods \cite{farha2019ms, huang2020gbtr, ishikawa2021dab, asformer, ltc2023bahrami} require expensive frame-wise labels, and weakly-supervised methods \cite{chang2019d3tw, ding2018weakly, kuehne2017weakly, li2019weakly, richard2017weakly, richard2018neuralnetwork, khan2022timestamp, li2021temporal, rahaman2022generalized, fayyaz2020sct, li2020set, richard2018action, souri2022robust, behrmann2022unified} still rely on prior knowledge of action classes. In order to discover semantically meaningful classes directly from long untrimmed videos without any labeling, unsupervised approaches for temporal action segmentation have received increasing attention \cite{kukleva2019unsupervised, vidalmata2021joint, li2021action, swetha2021unsupervised, kumar2022tot, tran2024ufsa, xu2024asot}.

\begin{figure}[tb]
    \centering
    \includegraphics[width=\linewidth]{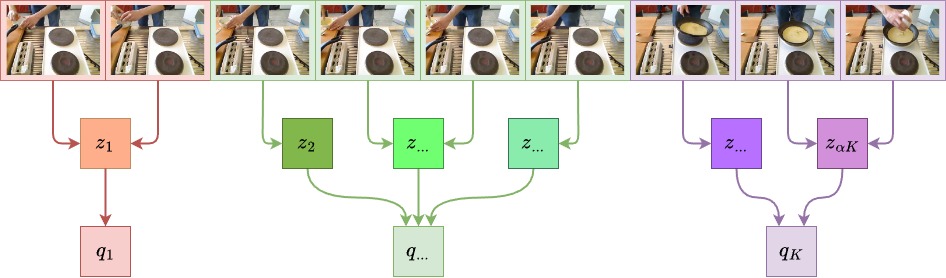}
    \caption{
    Given a set of long untrimmed video sequences (first row), the goal is to segment each video and cluster the segments across all videos. Our approach for hierarchical vector quantization learns two codebooks ${\bf Z}$ (second row) and ${\bf Q}$ (third row), which results in a hierarchical representation of actions.  
    }
    \label{fig:teaser}
\end{figure}

Early approaches for unsupervised action segmentation~\cite{kukleva2019unsupervised, vidalmata2021joint, li2021action} learn first a representation and then cluster and segment the frames based on the learned representation in a multi-step fashion. More recent approaches \cite{swetha2021unsupervised, kumar2022tot, tran2024ufsa, xu2024asot} combine representation learning and clustering in a single step, since representation learning affects clustering and vice-versa.
TOT~\cite{kumar2022tot}, for example, applies temporal optimal transport for generating pseudo-labels, which are then used to supervise the representation learning. The recent approach ASOT~\cite{xu2024asot} extends TOT by relaxing the constraint of knowing the action order for a video and it improves the performance. However, it introduces implicitly a strong prior on the segment length. It fails to identify very short segments, as shown in \figref{fig:qualitative_friedegg}, and the distribution over the segment lengths deviates largely from the ground-truth distribution, as shown in \figref{fig:count_distribution}. The bias is caused by generating pseudo-labels  from a noisy frame-to-cluster assignment matrix, which requires using strong priors like a structural prior. 

Since methods for temporal unsupervised action segmentation should not have a strong bias towards segments of certain lengths, we propose a new metric based on the Jensen-Shannon Distance (JSD), which measures such bias and complements existing metrics like the F1 score. To address this bias, we furthermore propose a novel approach that is based on a novel paradigm compared to previous approaches that either generate pseudo-labels or follow a multi-step approach. Our approach reformulates the problem as a vector quantization problem where a codebook is learned that represents the clusters. It learns an embedding of frames, a codebook and the assignment of frames to the codebook vectors, and thus clusters, jointly. In contrast to previous works, it neither requires the generation of pseudo-labels nor performs the embedding learning, clustering and assignment of frames in three separate steps.    

To deal with large intrinsic variations of semantically meaningful segments, we explicitly take in consideration the natural composition of actions, \ie, in order to complete an action it may be also necessary to complete intermediate steps. We thus propose a hierarchical vector quantization approach, termed Hierarchical Vector Quantization (\ours), that provides a fine-to-coarse clustering as illustrated in \figref{fig:teaser}. The quantization is performed at two levels. The first quantization deals with subactions, a fine-grained representation of an action, while the second quantization combines the subaction clusters to obtain the action representation. In this way, each coarse cluster can be represented by a varying number of finer clusters to cope with the intrinsic variation within each coarse cluster. \ours\ is the first hierarchical approach for unsupervised temporal action segmentation and the first approach that utilizes vector quantization.   

We evaluate the proposed approach on the Breakfast \cite{kuehne2014breakfast}, YouTube Instructional \cite{alayrac2016narrated} and IKEA ASM \cite{shabat2022IKEA} datasets. On all the datasets, the proposed approach achieves state-of-the-art results in terms of F1 score, recall, and the Jensen-Shannon Distance (JSD), which measures the similarity of the distributions of the segment lengths of the predicted and ground-truth segments.     
In summary, our contributions are:
\begin{itemize}
    \item We propose a novel end-to-end architecture for unsupervised temporal action segmentation. The hierarchical approach is able to capture the variability of action clusters and outperforms a non-hierarchical variant by a large margin.  
    \item We introduce a novel metric based on the Jensen-Shannon Distance for evaluating the bias in the predicted segment lengths. 
    \item Our method achieves state-of-the-art results for unsupervised temporal action segmentation in terms of recall and F1-score, and it is less biased. 
\end{itemize}
\section{Related Works}
\label{sec:related}

\paragraph{Fully-Supervised Action Segmentation.} Early fully-supervised action segmentation methods rely on a sliding temporal window and non-maximum suppression \cite{karaman2014, rohrbach2012} or model the problem with hidden Markov Models (HMM) \cite{kuehne2016end, tang2012latent}. \citet{richard2016temporal} model the probability of action sequences with language and length models. Recent methods mainly use temporal convolutional networks (TCNs) with temporal pooling to capture long-range dependencies, though this can result in some loss of fine-grained temporal information. For this reason, \citet{farha2019ms} introduced a multi-stage TCN, which maintains a high temporal resolution and has been further improved by \citet{li2020ms}. However, frame-wise predictions can lead to over-segmentation. To mitigate this, \citet{ishikawa2021dab} introduced a boundary regression branch, and \citet{huang2020gbtr} proposed a graph-based temporal reasoning module. With the rise of Transformers \cite{attention2017} in vision tasks \cite{vit2021, zheng2021rethinking, wang2021panoptic, arnab2021vivit, bertasius2021timesformer}, ASFormer \cite{asformer} combined multi-stage TCNs with transformers, but also faced over-segmentation. \citet{behrmann2022unified} adressed this by leveraging segment-level cues, while \citet{ltc2023bahrami} used sparse attention to capture a long temporal context. Lastly, \citet{liu2023diffusion} proposed a framework using denoising diffusion models with iterative refinement.

\paragraph{Weakly-Supervised Action Segmentation.} Weakly-supervised approaches aim to use a reduced level of supervision. These approaches vary based on the type of weak labels used. Transcript-supervised methods \cite{chang2019d3tw, ding2018weakly, kuehne2017weakly, li2019weakly, richard2017weakly, richard2018neuralnetwork, lu2021subspaces} know the action order in training videos, while set-supervised methods \cite{fayyaz2020sct, li2020set, richard2018action} only know the set of actions without their order or timing. Timestamp supervision \cite{khan2022timestamp, li2021temporal, rahaman2022generalized} requires the labeling of just one frame per action segment, offering improved results and similar annotation cost. \citet{souri2022robust} further improved this by handling missing annotations. Recently, \citet{behrmann2022unified} achieved competitive results by using a unified fully and timestamp-supervised method. Unlike these methods, our approach does not require any action labels.

\paragraph{Unsupervised Action Segmentation.} Two different frameworks have been explored for unsupervised action segmentation. The first framework achieves action segmentation by detecting boundary changes at video level, \ie\ processing and evaluating the videos individually \cite{du2022abs, aakur2019perceptual, sarfraz2021twfinch, mounir2023streamer}. The second framework, instead, processes and evaluates the videos at activity level, with an additional challenge of identifying the same actions across videos. Our work falls in this second framework.
Early works in unsupervised action segmentation utilise the narration for the videos for segmentation. This is possible only if the narration is provided. For this reason, recent methods only rely on visual features and exploit the temporal structure of videos. One of the early works, Mallow \cite{sener2018unsupervised}, alternates between learning a discriminative appearance model and optimizing a generative temporal model of the activity. CTE \cite{kukleva2019unsupervised} introduces a multi-step approach that starts from learning continuous temporal embeddings and performs a clustering on the learned features. JVT \cite{vidalmata2021joint} improves CTE by adding a visual embedding. TAEC \cite{taec} uses a encoder-decoder structure with a TCN for learning temporal embeddings, while ASAL \cite{li2021action} learns also an action level embedding. These methods use a multi-step approach, representation learning and then offline clustering. UDE~\cite{swetha2021unsupervised} and TOT~\cite{tran2024ufsa}, on the other hand, allow feedback between these steps, performing a joint representation learning and online clustering. UFSA~\cite{tran2024ufsa} follows these recent approaches, but instead of exploiting frame-level cues only, it uses also segment-level cues and pseudo-labels. The recent work ASOT~\cite{xu2024asot} relaxes the fixed action order constraint of TOT and UFSA, but it suffers from a strong bias on the lengths of estimated segments.     

\section{Hierarchical Vector Quantization for Unsupervised Action Segmentation}
\label{sec:method}

\begin{figure*}
  \centering
  \includegraphics[width=0.9\linewidth, scale=0.2]{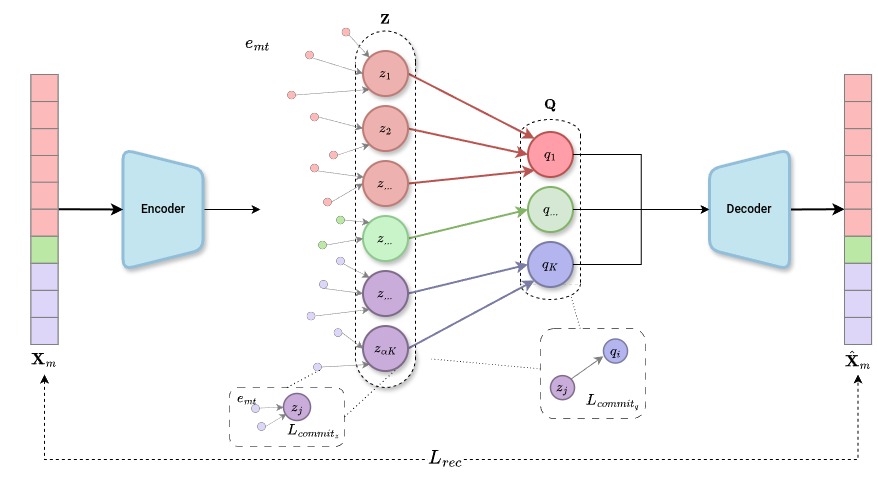}
  \caption{Overview of our model. The encoder takes as input a video $X_m$ and encodes every frame into an embedding vector $e_{mt}$. Every $e_{mt}$ is associated with the closest prototype $z_j \in {\bf Z}$. The prototypes $z_j$ are then associated with the closest prototype $q_i \in {\bf Q}$. The association of a frame to $\bf Q$ via ${\bf Z}$ defines the cluster assignment.}
  \label{fig:model}
\end{figure*}

Unsupervised temporal action segmentation is a challenging task since it requires to identify temporal clusters that correspond to semantically meaningful actions not only within a video, but also across videos, as shown in \figref{fig:qualitative_friedegg}. In particular, the cross-video setting imposes major challenges, since there is a large variation on how an action is executed by different persons and there is a large appearance variation if actions are recorded in different environments. Even though recent works based on generating pseudo-labels using optimal transport~\cite{kumar2022tot,tran2024ufsa,xu2024asot} have shown good performance, they introduce a segment length bias, which is particularly evident for the most recent approach ASOT~\cite{xu2024asot}, as shown in \figref{fig:count_distribution}. To quantify such bias, we will introduce a new metric based on the Jensen-Shannon Distance in \secref{sec:eval}. To address this bias, we propose a novel approach that is based on a novel paradigm compared to previous approaches that either generate pseudo-labels or follow a multi-step approach as discussed in \secref{sec:related}. Our approach reformulates the problem as a vector quantization problem where a codebook is learned that represents the clusters. In this way, we learn an embedding of frames, a codebook and the assignment of frames to the codebook vectors, and thus clusters, jointly. We demonstrate that the approach has a lower segment length bias compared to previous works. In order to obtain better clusters, we furthermore introduce a hierarchical approach that learns not one but two codebooks where each codebook represents a different granularity as illustrated in \figref{fig:teaser}.      

Specifically, our model consists of an auto-encoder structure where the hierarchy is composed by two-levels of vector quantization as illustrated in \figref{fig:model}. The encoder maps each frame $x_{mt}$ of a video $m$ with $T_m$ frames to an embedding vector $e_{mt}$. The first level of the vector quantizer then associates every encoded frame $e_{mt}$ with one embedding vector $\fvq \in {\bf Z}$ of the first learned codebook ${\bf Z}=\{z_1,...,z_{\alpha K}\}$, where $\alpha$ is a hyperparameter. The second stage vector quantizer associates every embedding vector $\fvq$ with another embedding vector $\svq \in {\bf Q}$ of the second learned codebook ${\bf Q}=\{q_1,...,q_K\}$. While the second codebook ${\bf Q}$ represents the $K$ clusters, the first codebook corresponds to more fine-grained clusters. Since the number of subactions that correspond to a semantically meaningful action may vary between actions, the number of embedding vectors $\fvq$ that are assigned to $\svq$ can vary. After the vector quantization, each frame is represented by $q_{mt} \in {\bf Q}$. The decoder then aims to reconstruct the input video sequence ${\bf X_m}=\{x_{mt}\}_{t=1}^{T_m}$ from $\{q_{mt}\}_{t=1}^{T_m}$. We denote the reconstruction by ${\bf \hat{X}_m} = \{\hat{x}_{mt}\}_{t=1}^{T_m}$ and the codebooks ${\bf Z}$ and ${\bf Q}$ are learned by minimizing the reconstruction error. We now describe the method in detail.

\subsection{Model Details} 
For the encoder and decoder, we use a light version of MS-TCN \cite{farha2019ms}. The encoder and decoder have the same structure. They both consist of two stages, where each stage is composed by $10$ layers. Each layer includes a temporal convolution with a dilation factor, which is doubled at each layer. Dilated temporal convolutions are very useful for the decoder since after quantization many neighboring frames are represented by the same vector $\svq$ and the reconstruction requires a large receptive field. We evaluate the impact of the TCN in the GitHub repository.

The prototypes are updated dynamically as a new video is fed to the network, making our clustering online. Given the current codebook ${\bf Z}$, each frame is associated to the closest prototype $z_j \in {\bf Z}$:  
\begin{align}
z_{mt} = z_{\hat{j}}, \quad \hat{j} = \argmax_j \frac{\fvq \cdot e_{mt}}{\|\fvq\| \|e_{mt}\|}.
\end{align}
The association of a subcluster to a cluster is done in the same way by   
\begin{align}
q_{mt} = q_{\hat{i}}, \quad \hat{i} = \argmax_i \frac{\svq \cdot z_{mt}}{\|\svq\| \|z_{mt}\|},
\end{align}
where the association needs only to be computed between $\fvq \in {\bf Z}$ and $\svq \in {\bf Q}$ and not all frames. As in \cite{oord2017vqvae}, we update the prototypes in the codebooks by the exponential moving average with $\beta=0.8$:
\begin{align}
    &\hat{z}_j = \frac{1}{\hat{N}_{z_j}}\left(\beta z_j +  (1-\beta)\sum_{z_{mt}=z_j} e_{mt}\right) \\
    \nonumber&\text{where } \hat{N}_{z_j} = \beta N_{z_j} + (1-\beta) \vert\{z_{mt}=z_j\}\vert, \\[10pt]
    &\hat{q}_i = \frac{1}{\hat{N}_{q_i}}\left(\beta q_i +  (1-\beta)\sum_{q_{mt}=q_i} z_{mt}\right) \\
    \nonumber&\text{where } \hat{N}_{q_i} = \beta N_{q_i} + (1-\beta) \vert\{q_{mt}=q_i\}\vert.
\label{eq:ema}
\end{align}
${N}_{z_j}$ is the previous estimation of $\hat{N}_{z_j}$, which represent the number of assigned vectors to prototype $z_j$, and similarly ${N}_{q_j}$.

In addition, we also use the strategy proposed by \citet{dhariwal2020jukebox}. When a prototype has not been assigned to any input for several batches, we replace the prototype with an input vector randomly sampled within the current batch. 
Specifically, we apply this when $\hat{N}_{z_j}<3$ and $\hat{N}_{q_i}<1$.
Furthermore, we apply $l_2$ normalization on the encoded latent variables $e_{mt}$ and codebook latent variables $\fvq$ to improve training stability and reconstruction quality \cite{yu2022vqgan}.

\subsection{Training} We train our model using the reconstruction loss. For a video $m$, it is given by
\begin{equation}
  L_{rec} = \sum_{t=1}^{T_m} \lvert\lvert x_{mt} - \hat{x}_{mt} \rvert\rvert_2^2.
  \label{eq:rec-loss}
\end{equation}
In addition, we use a commitment loss \cite{oord2017vqvae} which encourages the encoding vectors to stay close to the chosen prototype and to prevent frequent fluctuations in the assignment of code vectors~\cite{razavi2019vqvae2}: 
\begin{equation}    
\begin{aligned}    
  L_{commit_z} &= \sum_{t=1}^{T_m}\|e_{mt} - \text{sg}[z_{mt}]\|^2_2, \\
  L_{commit_q} &= \sum_{t=1}^{T_m}\|z_{mt} - \text{sg}[q_{mt}]\|^2_2,
  \label{eq:fvq-loss}
\end{aligned}
\end{equation}
where $\text{sg}[\cdot]$ denotes stop-gradient. In this way, the embeddings are pushed towards the closest prototype and not the other way around.   

The overall loss for training our model is then the combination of the commitment loss for the vector quantization and the reconstruction loss of the auto-encoder:
\begin{equation}
    L =  L_{commit_z} + L_{commit_q} + \lambda_{rec} \cdot L_{rec}
    \label{eq:final-loss}
\end{equation}
where $\lambda_{rec}$ weights the two loss terms. We provide additional implementation details in the supp.\ material, which also includes additional ablation studies on the impact of the hyperparameters.  

\subsection{Inference}
\label{sec:inference}
After having trained the model, we obtain for each video $m$ and each frame $x_{mt}$ an assignment to a prototype $q_{mt} \in \{q_1,...,q_K\}$, where $q_i$ is a prototype for cluster $i \in \{1,...,K\}$. In order to obtain a smooth solution over time, we apply FIFA~\cite{fifa2021}. To this end, we convert the hard assignment of a frame to a cluster into a soft assignment. This is achieved by computing for a frame $x_{mt}$ the distances between the encoding $e_{mt}$ and the prototypes in $z_j \in {\bf Z}$ and the distances between prototypes $z_j \in {\bf Z}$ and $q_i \in {\bf Q}$:  
\begin{equation}
    sim_{mt,i} = \sum_{j=1}^{\vert \bf Z\vert} (1-d(e_{mt},z_j)) + (1-d(z_j,q_i))
    \label{eq:dist}
\end{equation}
where $d(\cdot,\cdot)$ denotes the cosine distance. Using the softmax over $sim_{mt,i}$ then gives the soft assignment of frame $x_{mt}$ to cluster $i$.    
Besides the assignment likelihood, FIFA takes a cluster order as input. As in \cite{kukleva2019unsupervised}, we compute the mean over the timestamps of all frames belonging to each cluster and order the clusters with respect to this average timestamp. For the length prior, we count how often a cluster has been assigned to a frame in a video and use the average percentage over all videos.  

\section{Experiments}
\label{sec:exp}

\subsection{Datasets} We evaluate the proposed method on three public datasets: Breakfast \cite{kuehne2014breakfast}, YouTube Instructional \cite{alayrac2016narrated} and IKEA ASM \cite{shabat2022IKEA}. 
\begin{itemize}
    \item The Breakfast dataset is a large-scale dataset ($1712$ videos) that includes $70$ hours of videos of $10$ different common kitchen activities. The duration of videos varies widely from $30$ seconds to few minutes. It has a total of $48$ subaction classes.   
    \item The YouTube Instructional (YTI) dataset, the smallest used in this work, contains $150$ videos from YouTube with an average length of $2$ minutes per video. It captures $5$ activities with $47$ subaction classes. This dataset contains a background class and the fraction of it varies from $46\%$ to $83\%$. 
    \item The IKEA ASM (IKEA) dataset contains $371$ videos recording people assembling IKEA furniture for a total of $35$ hours. There are $4$ activities with $33$ actions. The length of the videos varies from $1$ to $8$ minutes approximately. 
\end{itemize}
As input, we use Improved Dense Trajectory (IDT) features \cite{wang2013idt} for the Breakfast dataset and the features provided by \citet{alayrac2016narrated} for the YouTube Instructional dataset. For the IKEA dataset, we have extracted the DINO features \cite{caron2021dino}. 

\subsection{Evaluation Protocol} 
\label{sec:eval}
Following the protocol that has been introduced by \citet{kukleva2019unsupervised}, we apply our approach to all videos of each activity separately. $K$ is set to the max number of subactions that appear for each activity. This is required for a fair comparison of the methods. We establish the mapping between predicted cluster segments and ground-truth segments via Hungarian matching, which is computed over all videos and their frames of one activity. 

For all the datasets, we report four metrics: the Mean over Frames (MoF), the F1-score~\cite{kukleva2019unsupervised}, precision and recall. The MoF is the percentage of frames with correct predictions averaged over all activities. MoF is not a reliable measure for imbalanced datasets because it is dominated by frequent and long action classes, whereas the impact of short segments, which are much more difficult to detect, is very low on MoF. MoF thus prefers methods with a bias towards longer segments.     
The F1-score~\cite{kukleva2019unsupervised}, which is the harmonic mean between precision and recall, is a much better metric. 
In addition to the F1 score, we also report recall, which measures how many segments are missed. 

A key aspect of this work is to analyze if methods have a bias in terms of the length of the generated segments. To quantify this, we propose a new metric based on the Jensen-Shannon Distance (JSD). For each video within the same activity, we compute the histogram of the predicted segment lengths, using a bin width of 20 frames. We then compare this histogram with the corresponding ground-truth histogram using the Jensen-Shannon Distance. These JSD scores are averaged across all videos for each activity. Finally, we calculate a weighted average across all activity, where the weights are the number of frames in each activity. In particular:
\begin{equation}
    \text{JSD} = \frac{\sum_{a \in A} F_a \cdot \frac{1}{|M_a|} \sum_{m \in M_a} \text{JSDist}(H_m^{\text{pred}}, H_m^{\text{gt}})}{\sum_{a \in A} F_a},
\label{eq:jsd-formula}
\end{equation}
where $F_a{=}\sum_{m=1}^{M_a} T_m$ is the total number of frames and  $M_a$ is the set of all the videos for activity $a \in A$. JSDist is the Jensen-Shannon Distance:
\begin{equation}
    \text{JSDist}(P, Q) = \sqrt{\frac{D_{\text{KL}}(P \parallel M) + D_{\text{KL}}(Q \parallel M)}{2}},
\label{eq:jsd-classic}
\end{equation}
where $M{=}\frac{P+Q}{2}$ and $D_{\text{KL}}$ is the Kullback-Leibler divergence. 
The input of JSDist is normalized such that the sum of each histogram $H_m$ is one. 
The metric measures how dissimilar the distributions of the lengths of the predicted and ground-truth segments are. A high value indicates a strong bias in segment lengths. Note that we do not report the metric for YTI since the video sequences are masked based on the ground-truth background for evaluation, but we report it for the other datasets that are evaluated without masking.  

\begin{table}[tb]
    \centering
    \footnotesize
    \setlength{\tabcolsep}{2pt}
    \resizebox{\columnwidth}{!}{%
        \begin{tabular}{c@{\hspace{0.5em}}ccccc}
            \toprule
            \textbf{Method}  
            & \textbf{MOF} & \textbf{F1} & \textbf{Recall*} & \textbf{JSD*$\downarrow$} & {\bf Precision*} \\
            \midrule
            
            Mallow~\cite{sener2018unsupervised} 
            & 34.6 & - & - & - & - \\            
            CTE~\cite{kukleva2019unsupervised} 
            & 41.8 & 26.4 & 27.0 & 87.4 & 25.8 \\
            JVT~\cite{vidalmata2021joint} 
            & 48.1 & - & - & - & - \\
            ASAL~\cite{li2021action} 
            & 52.5 & 37.9 & - & - & - \\
            UDE\textdagger~\cite{swetha2021unsupervised} 
            & 47.4 & 31.9 & - & - & - \\
            TOT~\cite{kumar2022tot} 
            & 47.5 & 31.0 & 26.3 & 90.2 & {\bf 37.7} \\
            TOT+TCL~\cite{kumar2022tot} 
            & 39.0 & 30.3 & 36.0 & \underline{\it 85.6} & 26.2 \\
            UFSA~\cite{tran2024ufsa} 
            & 52.1 & 38.0 & - & - & - \\
            ASOT~\cite{xu2024asot} 
            & {\bf 56.1} & \underline{\it 38.3} & \underline{\it 40.1} & 94.9 & \underline{\it 36.7} \\
            Ours (\ours) 
            & \underline{\it \bfm} & {\bf \bff} & {\bf \bfr} & {\bf \bfjs} & 35.6 \\
            
            \bottomrule        
        \end{tabular}  
    }
    \caption{Results on Breakfast. Best results are in \textbf{bold}, while second best ones are \underline{\textit{underlined}}. \textdagger denotes results based on I3D features instead of IDT. * denotes metric computed by ourselves.}
    \label{tab:breakfast_results}    
\end{table}

\begin{figure*}[t]
    \centering
    \includegraphics[width=1.0\linewidth]{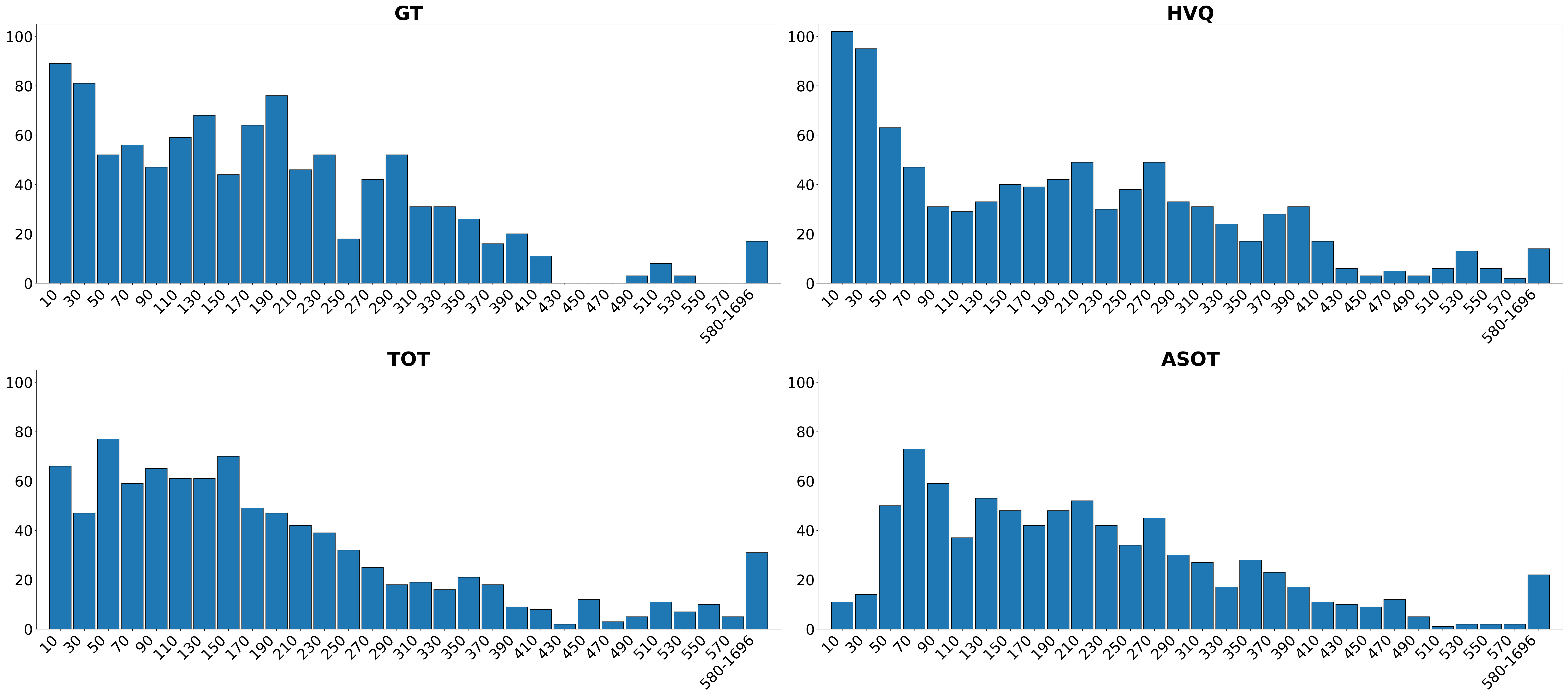}
    \caption{Histograms of the segment lengths across the whole milk activity
    for ground-truth, \ours, TOT and ASOT. Segments longer than $580$ frames are in the last bin.}
    \label{fig:count_distribution}
\end{figure*}

\subsection{Comparison to State of the Art}\label{sec:comp}

\paragraph{\bf Result on Breakfast.} 
Table \ref{tab:breakfast_results} presents the results of our method compared to state-of-the-art approaches for unsupervised temporal action segmentation. Our approach outperforms the state of the art in terms of F1 score, recall and JSD, and it achieves the second best MoF. Only the recent work ASOT~\cite{xu2024asot} achieves a higher MoF. Our approach, however, outperforms ASOT for all the other metrics except of precision. For instance, it achieves a higher recall, \ie, it misses less action segments than ASOT. Furthermore, ASOT has also the highest JSD indicating the strongest bias of segment lengths compared to other approaches. The relatively high MoF of ASOT compared to F1 score, recall and JSD can be explained by a bias of ASOT to longer segments, while missing short ground-truth segments that have a very low impact on MoF. 

To analyze this further, \figref{fig:count_distribution} plots the histograms of the segment lengths across the whole milk activity, using a bin width of $20$ frames. The plots show the strong length bias of ASOT. While there are many short segments (first two bins) in the ground-truth, ASOT generates only very few short segments. These segments have a very low impact on MoF, in contrast to the other metrics, and explain the high MoF of ASOT. TOT \cite{kumar2022tot} is less biased as shown by the histograms and JSD, but short segments are underrepresented as well. Our approach follows the ground-truth distribution of segment lengths best, and thus achieves the lowest JSD. While TOT+TCL achieves the second best JSD, the segments are often misclassified, which results in a low F1-score and recall in Table \ref{tab:breakfast_results}. Our approach achieves $+9.4$ and $+8.9$ higher F1-score and recall than TOT+TCL, respectively.

\begin{figure}[tb]
    \centering
    \includegraphics[width=1.0\linewidth]{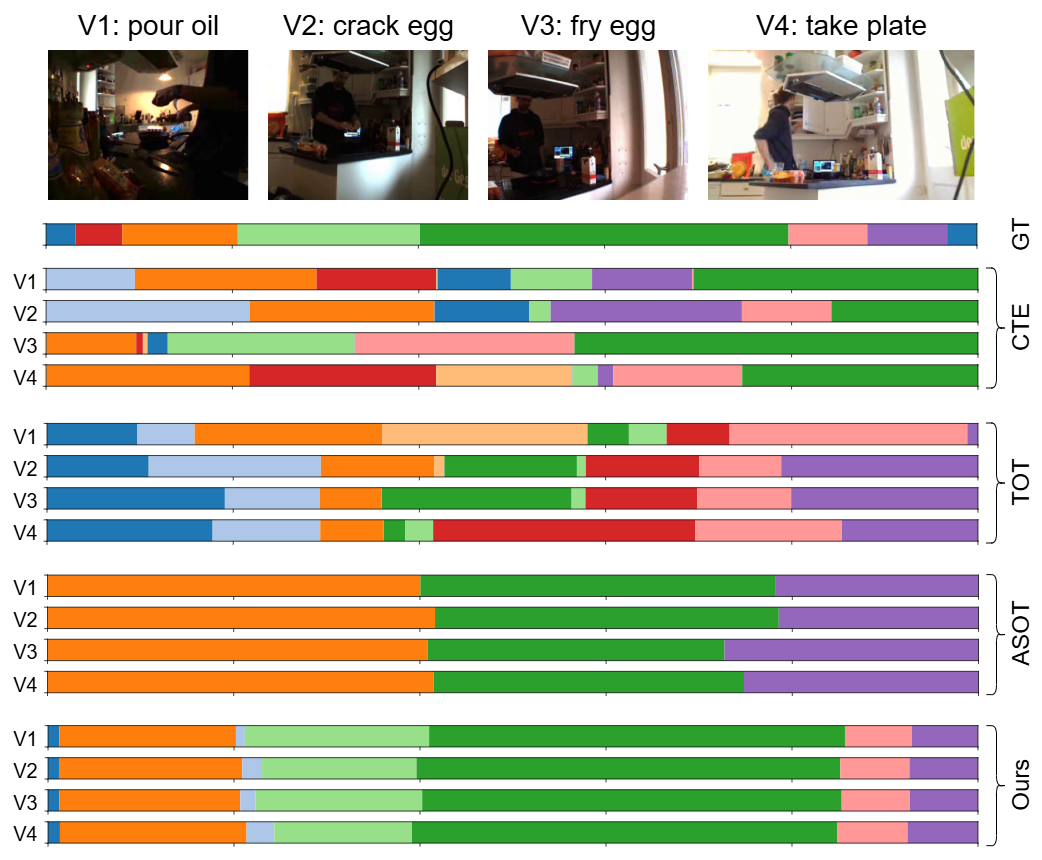}
    \caption{Segmentation results for a sample from the Breakfast dataset ({\it P22\_friedegg}). 
    Our approach delivers highly consistent results across multiple videos (V1, V2, V3, V4) recorded from different cameras, but with the same ground truth.
    }
    \label{fig:qualitative_friedegg}
\end{figure}

\figref{fig:qualitative_friedegg} shows qualitative segmentation results for a video from the Breakfast dataset. The activity is \textit{friedegg} and it contains short and long actions. We compare to CTE \cite{kukleva2019unsupervised}, TOT \cite{kumar2022tot} and ASOT~\cite{xu2024asot}. V1, V2, V3 and V4 represent four different videos of the same sequence ("{\it P22\_friedegg}") recorded with different cameras, thus the ground-truth segmentation is the same for all the four videos. 
For CTE and TOT, the predictions are very different across videos, therefore the predictions are not consistent even though the sequence is the same. 
In contrast, the predictions of ASOT do not change much across videos, but the network assigns the first four actions to a unique cluster, and the same for the last three actions. As discussed, ASOT is strongly biased towards longer segments, often overlooking shorter ones.
Our method is not only consistent across V1, V2, V3 and V4, but it is also capable of correctly clustering the long dark green action and the shorter actions. 
However, our approach fails as the other methods to identify the small red segment of {\it pour oil}. 
   
\begin{table}[tb]        
    \centering
    \footnotesize
    \resizebox{\columnwidth}{!}{%
        \begin{tabular}{c@{\hspace{1em}}cccc}
            \toprule
            \textbf{Method} 
            & \textbf{MOF} & \textbf{F1} & \textbf{Recall*} & {\bf Precision*} \\
            \midrule        
            Frank-Wolfe~\cite{alayrac2016narrated} 
            & - & 24.4 & - & - \\   
            Mallow~\cite{sener2018unsupervised} 
            & 27.8 & 27.0 & - & - \\   
            CTE~\cite{kukleva2019unsupervised}
            & 39.0 & 28.3 & 22.1 & 39.3\\
            VTE~\cite{vidalmata2021joint} 
            & - & 29.9 & - & - \\
            ASAL~\cite{li2021action} 
            & 44.9 & 32.1 & - & - \\
            UDE\textdagger~\cite{swetha2021unsupervised} 
            & 43.8 & 29.6 & - & - \\
            TOT~\cite{kumar2022tot} 
            & 40.6 & 30.0 & \underline{\it 31.4} & 28.7 \\
            TOT+TCL~\cite{kumar2022tot} 
            & 45.3 & \underline{\it 32.9} & 27.9 & \underline{\it 40.1} \\
            UFSA~\cite{tran2024ufsa} 
            & \underline{\it 49.6} & 32.4 & - & - \\
            ASOT~\cite{xu2024asot}
            & {\bf 52.9} & {\bf 35.1} & 27.8 & {\bf 47.6} \\
            Ours (\ours) & {\ytim} &  {\bf \ytif} & {\bf \ytir} & 32.1 \\        
            \bottomrule        
        \end{tabular}   
    }
    \caption{Results on YouTube Instructions. Best results are in \textbf{bold}, while second best ones are \underline{\textit{underlined}}. \textdagger denotes results based on I3D features instead of IDT. * denotes metric computed by ourselves.}
    \label{tab:yti_results}
\end{table}

\begin{table}[tb]        
    \centering
    \footnotesize
    \setlength{\tabcolsep}{2pt}
    \resizebox{\columnwidth}{!}{%
        \begin{tabular}{c@{\hspace{0.5em}}ccccc}
            \toprule
            \textbf{Method} &
            \textbf{MOF} & \textbf{F1} & \textbf{Recall*} & \textbf{JSD*$\downarrow$} & {\bf Precision*} \\
            \midrule
            
            CTE*~\cite{kukleva2019unsupervised}
            & 23.1 & 22.6 & 18.9 & \underline{\it 73.7} & \underline{\it 28.1} \\
            TOT*~\cite{kumar2022tot} 
            & 21.0 & 20.1 & 17.1 & 80.0 & 24.4 \\
            TOT+TCL*~\cite{kumar2022tot} 
            & 23.8 & 20.9 & 17.7 & 79.5 & 25.5 \\
            ASOT*~\cite{xu2024asot}
            & \underline{\it 34.0} & \underline{\it 27.9} & \underline{\it 24.0} & 88.7 & 21.1 \\
            Ours (\ours)
            & {\bf \ikeatm} & {\bf \ikeatf} & {\bf 25.9} & {\bf \ikeajs} & {\bf 37.7} \\
            
            \bottomrule        
        \end{tabular}   
    }
    \caption{Results on IKEA ASM. Best results are in \textbf{bold}, while second best ones are \underline{\textit{underlined}}. * denotes metric computed by ourselves.}
    \label{tab:ikea_results}
\end{table}

\paragraph{\bf Results on YouTube Instructional.}
In \tabref{tab:yti_results}, we report the comparison between our method and the state of the art on the YouTube Instructional (YTI) dataset. We follow the protocol of prior works and evaluate the results excluding background frames. As for Breakfast, we outperform UDE \cite{swetha2021unsupervised} and TOT \cite{kumar2022tot} by a large margin. While the temporal coherence loss (TCL) decreases MoF and F1-score of TOT on Breakfast, it improves the results on YTI. Nevertheless, our approach outperforms TOT+TLC on both datasets. Our approach also outperforms UFSA \cite{tran2024ufsa}. As on Breakfast, ASOT achieves higher MOF and precision, but lower recall. The F1-scores of our method and ASOT are the same.

\paragraph{\bf Results on IKEA ASM.}
In \tabref{tab:ikea_results}, we report our results compared to the state-of-the-art approaches on the IKEA ASM dataset. We compare our approach to CTE, TOT, considering also the version with temporal consistency loss TOT+TCL, and ASOT. Our approach outperforms all methods in all metrics and the predicted segment lengths are closest to the ground-truth as measured by JSD.  

\subsection{Ablation Studies}
\label{sec:ablations}

\begin{table}[tb]
    \centering
    \footnotesize
    \begin{tabular}{@{}lccc@{}}
        \toprule
        &\textbf{MOF} & \textbf{F1} & \textbf{JSD} $\downarrow$ \\
        \midrule
        $L_{rec}$ & 41.6 & 29.7 & 87.2\\
        \midrule
        $L_{commit_q}$ & 46.0 & 35.1 & 84.4 \\
        $L_{commit_z}$ & 51.8 & 37.4 & 83.6 \\
        $L_{commit_z} + L_{commit_q}$ & {\bf 54.4} & 38.0 & {\bf 82.5} \\
        \midrule
        $L_{rec} + L_{commit_q}$ & 47.8 & 36.4 & 84.1 \\
        $L_{rec} + L_{commit_z}$ & 51.3 & 38.4 & 83.7 \\
        $L_{rec} + L_{commit_z} + L_{commit_q}$ & {\bf \bfm} & {\bf \bff} & {\bf \bfjs} \\
        \bottomrule    
    \end{tabular}    
    \caption{Impact of loss terms on the Breakfast dataset. Best results are in \textbf{bold}.}
    \label{tab:losses}
\end{table}

\paragraph{\bf Impact of Loss Terms.}
We report the impact of the loss terms for the Breakfast dataset in \tabref{tab:losses}. Using only the reconstruction loss results in a relatively low performance, which is expected since the reconstruction loss does not enforce compact clusters. This is achieved by the commitment loss. Using the commitment loss for both levels of the hierarchy $L_{commit_z} + L_{commit_q}$ already performs well. From the table, we can also see that $L_{commit_z}$ has a larger impact than $L_{commit_q}$, which is expected since $L_{commit_z}$ maps the frame-wise embeddings $e_{mt}$ to the fine-grained clusters whereas $L_{commit_q}$ associates the fine-grained clusters to the coarse clusters. Adding the reconstruction loss to the commitment loss improves the F1 score further.         

\paragraph{\bf Impact of $\lambda_{rec}$.} We also study the impact of the weight parameter $\lambda_{rec}$ for the loss term \eqref{eq:final-loss} and we report the results in \tabref{tab:rec_lambda}. Using $\lambda_{rec}=0.002$ performs well and we use it for all other experiments. The parameter is not sensitive to the dataset.    

\begin{table}[tb]
    \centering
    \footnotesize
    \begin{tabular}{cccccc}
        \toprule
         $\lambda_{rec}$ & 0.0005 & 0.001 & 0.002 & 0.005 & 0.01 \\
         \midrule
         Breakfast & 36.4 & 38.8 & {\bf \bff} & 38.9 & 37.8 \\
         YTI & 33.2 & 33.0 & {\bf \ytif} & 34.1 & 33.2 \\
         IKEA & 30.0 & 29.6 & {\bf \ikeatf} & 29.1 & 30.3 \\
         \bottomrule
    \end{tabular}
    \caption{F1 on Breakfast and YTI for different values of $\lambda_{rec}$}
    \label{tab:rec_lambda}
\end{table}

\paragraph{\bf Impact of $\alpha$.}
In \tabref{tab:alpha}, we show the results on the Breakfast dataset considering different numbers of prototypes in ${\bf Z}$, which is steered by different values of $\alpha$. While there is an improvement for $\alpha>1$ compared to $\alpha=1$, the performance saturates at $\alpha=2$. For larger values of $\alpha$, the performance decreases except for JSD since the subclusters become too fine-grained.  

\begin{table}[tb]     
    \centering
    \footnotesize
    \begin{tabular}{@{}l@{\hspace{2em}}cccc@{}}
        \toprule
        & $\alpha=1$ & $\alpha=2$ & $\alpha=3$ & $\alpha=4$\\
        \midrule
        {\it MOF} & 53.6 & {\bf \bfm} & 52.7 & 51.8 \\
        {\it F1} & 38.2 & {\bf \bff} & 38.3 & 38.2 \\        
        {\it JSD $\downarrow$} & 83.7 & \bfjs & 83.0 & {\bf 82.2} \\        
        \bottomrule
    \end{tabular}    
    \caption{Impact of $\alpha$ on the Breakfast dataset. Best results are in \textbf{bold}.}
    \label{tab:alpha}
\end{table}

\begin{table}[tb]
    \centering
    \footnotesize
    \begin{tabular}{@{}lcccc@{}}
        \toprule
        Dataset & Metric & Single & Double & Triple \\
        \midrule        
        YTI & F1 & 33.0  & {\bf \ytif} & 31.9 \\
        \midrule
        \multirow{2}{*}{IKEA ASM} 
        & F1 & 25.8 & {\ikeaf} & {\bf \ikeatf} \\
        & JSD & 81.9 & {\bf 62.2} & \ikeajs \\
        \midrule
        \multirow{2}{*}{Breakfast} 
        & F1 & 37.1 & {\bf \bff} & 38.2 \\        
        & JSD & 83.1 & {\bf \bfjs} & 84.1 \\            
        \midrule
        BF short & F1 & 38.3 & {\bf 39.9} & 39.0 \\
        BF medium & F1 & 41.4 & {\bf 42.5} & 40.0 \\
        BF long & F1 & 35.7 & {\bf 38.4} & 38.0 \\   
        \bottomrule        
    \end{tabular}
    \caption{Impact of the number of hierarchy levels.}
    \label{tab:levels}  
\end{table}

\paragraph{\bf Impact of the Hierarchy.} To show the benefits of having coarse and fine clustering, we report in \tabref{tab:levels} a comparison between single vector quantization and hierarchical vector quantization, with two and three levels. The results show that the hierarchical vector quantization with two levels outperforms the vector quantization with one level by a large margin in terms of F1 score. JSD is also better in all cases where the biggest improvement can be observed for IKEA ASM. Adding a third layer to the hierarchy does not improve the results, except for the F1 score on IKEA ASM.     
Compared to Breakfast and YTI, IKEA ASM has the highest average action length. To analyze the relation of hierarchy level and average action length more in detail, we group the $10$ activities of Breakfast into three sets, based on their average action segment lengths. Activities with short actions ({\it coffee, tea, cereals}) have an average action length below $150$ frames. Activities with medium-range actions ({\it sandwich, juice, milk}) have average action length between $150$ and $250$ frames. Finally, activities with long action segments ({\it fried-egg, salat, pancake, scrambled-egg}) have average action length above $250$ frames. The results in \tabref{tab:levels} show that the difference between having two or three layers is smallest for the activities with long average length. While using more than two layers depends on the expected average segment lengths, using a hierarchical approach with 2 layers always outperforms a non-hierarchical approach.   

In \tabref{tab:times}, we report the runtime of our method without hierarchy (single), with hierarchy (double), and with a large value of $\alpha$, and compare it to TOT, CTE and ASOT. We measured the runtime on the same workstation with Intel i9-13900k CPU with 24 cores and one NVIDIA RTX 3090 GPU. The overhead added by the hierarchy or by a larger codebook size ($\alpha=5$) is negligible. The runtime is the same as ASOT, but lower than CTE and TOT. We expect that the runtime of UFSA~\cite{tran2024ufsa} is higher than the runtime of TOT, since UFSA extends TOT by additional steps. The reported runtime of ASAL~\cite{li2021action} is roughly $10$ hours on Breakfast.

Additional ablation studies are provided in the supp.\ material.

\begin{table}[tb]
    \centering
    \footnotesize
    \begin{tabular}{@{}l@{\hspace{2em}}l@{}}
        \toprule
        \multicolumn{2}{c}{\bf Breakfast} \\
        \midrule
        {\bf Model} & \textbf{Time (minutes)}  \\
        \midrule
        TOT & 43 \\
        CTE & 23 \\
        ASOT & 18 \\
        Ours (Single) & 18:48 \\    
        Ours (Double, $\alpha=2$) & 19:04 \\
        Ours (Double, $\alpha=5$) & 19:15 \\
        \bottomrule
    \end{tabular}    
    \caption{Training and inference time for processing the entire Breakfast dataset. }
    \label{tab:times}
\end{table} 

\section{Conclusion}
In this paper, we introduced Hierarchical Vector Quantization (\ours), a novel paradigm for the problem of unsupervised temporal action segmentation. While the first level of the hierarchy captures fine-grained subactions, the second one refines these predictions and produces the desired action-level granularity. Furthermore, we introduced a new metric that measures the bias in the generated segment lengths. We showed that our approach is less biased than previous works and that it
is capable of dealing with the large variations of temporal segments.  
Additionally, the resulting vector quantization clusters are consistently assigned to the same actions across different videos. 
We provided an extensive set of ablations studies to show the efficiency of the proposed approach, and showed that our method achieves state-of-the-art results over previous methods on the Breakfast, YouTube Instructional and IKEA ASM datasets.

\section*{Acknowledgments}
The work has been supported by the Deutsche Forschungsgemeinschaft (DFG, German Research Foundation) GA 1927/4-2 (FOR 2535 Anticipating Human Behavior), the project iBehave (receiving funding from the programme “Netzwerke 2021”, an initiative of the Ministry of Culture and Science of the State of Northrhine Westphalia), and the ERC Consolidator Grant FORHUE (101044724). 
The sole responsibility for the content of this publication lies with the authors.

\bibliography{aaai25}

\clearpage

\appendix

\section{Implementation Details} 
We train our model end-to-end with AdamW as optimizer, with a learning rate of $10^{-3}$ and a weight decay of $10^{-4}$. 
The encoder and decoder architectures are two-stage MS-TCNs with $10$ layers for each stage, where we removed the classification head of the last stage. 
For YouTube Instructional, we use dropout after each layer of MS-TCN due to the large amount of background present in the dataset. The dimension $D$ of the encoding vectors in the latent space is set to $32$. For Breakfast and YouTube Instructional, which contain both activities with short and medium/long action segments, we use two levels of hierarchy; for IKEA, instead, which contains only activities with medium/long actions, we use three levels of hierarchy. We set $\alpha$ to $2$ for Breakfast and $3$ for YouTube Instructional due to the background. For IKEA, we set $\alpha$ to $2$ for the first two levels of the hierarchy. If not otherwise specified, we use $\lambda_{rec} = 0.002$. For inference, we use FIFA decoding with a learning rate of $6 \times 10^{-6}$, sharpness of $0.1$ and $100$ epochs.

\subsection{\bf Encoder and Decoder}
We use a light version of MS-TCN with 2 stages as encoder and decoder. In \tabref{tab:ms-stages}, we evaluate the impact of the number of stages for the encoder and decoder. Using two stages performs better than using only one stage. Increasing the number of stages further decreases the accuracy. We assume that in the unsupervised case the encoder and decoder are not anymore properly trained when they have too many parameters. In \tabref{tab:decoder}, we also compare the MS-TCN decoder to an MLP decoder. The MLP decoder does not consider temporal information for decoding, which results in a lower performance. In the last row of \tabref{tab:decoder}, we evaluate the impact of FIFA decoding for inference, which smooths the predictions.       

\begin{table}[b]
    \centering
    \footnotesize
    \begin{tabular}{@{}l@{\hspace{2em}}cccc@{}}
        \toprule
        & 1 stage & 2 stages & 3 stages & 4 stages \\
        \midrule        
        \textit{MOF} & 51.8 & {\bf \bfm} & 46.4 & 47.5 \\
        \textit{F1} &  34.4 & {\bf \bff} & 35.9 & 36.9 \\ 
        \bottomrule
    \end{tabular}    
    \caption{Impact of number of stages on MS-TCN encoder and decoder on the Breakfast dataset. Best results are in \textbf{bold}.}
    \label{tab:ms-stages}
\end{table}

\begin{table}[b]
    \centering
    \footnotesize
    \begin{tabular}{@{}l@{\hspace{2em}}|cc|c@{}}
        \toprule
        & MLP & MS-TCN & Ours w/o FIFA \\
        \midrule
        {\it MOF} & 50.9 & {\bf \bfm} & 48.6 \\
        {\it F1} & 37.5 & {\bf \bff} & 29.9 \\
        \bottomrule    
    \end{tabular}       
    \caption{Different decoder networks on the Breakfast dataset. Best results are in \textbf{bold}.}
    \label{tab:decoder}
\end{table}

\subsection{\bf Importance of Vector Quantization.} In \tabref{tab:kmean}, we evaluate the impact of vector quantization. If we use a two step approach, \ie, we first train our auto-encoder with the reconstruction loss and then apply K-Means clustering on the encoded features, we obtain $26.4$ MoF and $25.4$ F1-score. This shows that the performance is mainly due to the proposed vector quantization and not due to the auto-encoder. 
We also evaluate the impact of initializing the prototype vectors using K-Means. We apply K-Means on the first training video and use the learned centroids as initialization. If we initialize the prototypes by random vectors, the performance is lower with $47.9$ MoF and $37.5$ F1.

\begin{table}[tb]
    \centering
    \footnotesize
    \begin{tabular}{@{}l@{\hspace{2em}}ccc@{}}
        \toprule
        & K-Means & Ours w/ random init. & Ours \\
        \midrule
        {\it MOF} & 26.4 & 47.9 & {\bf \bfm} \\
        {\it F1} & 25.4 & 37.5 & {\bf \bff} \\
        \bottomrule  
    \end{tabular}    
    \caption{Comparison to K-Means and random initialization for Breakfast. Best results are in \textbf{bold}.}
    \label{tab:kmean}
\end{table}

\subsection{Impact of decay factor $\beta$}
In \tabref{tab:decay}, we evaluate the impact of the decay factor $\beta$ that regulates the Exponential Moving Average (EMA) for updating the prototypes.

\begin{table}[tb]
    \caption{F1-score on Breakfast with different values of Exponential Moving Average (EMA) decay factor.}
    \centering
    \footnotesize
    \begin{tabular}{cccccc}
        \toprule
         {\bf EMA Decay} & 0.7 & 0.75 & 0.8 & 0.85 & 0.9  \\
         \midrule
         {\it F1} & 38.3 & 37.9 & {\bf \bff} & 39.2 & 38.4 \\ 
         \bottomrule
    \end{tabular}
    \label{tab:decay}
\end{table}

\section{Additional Comparisons}

\subsection{Additional protocol}
Following \cite{kumar2022tot, tran2024ufsa}, we evaluate our approach using a second protocol. We train the model with $80\%$ of the videos in the dataset and test it on the remaining $20\%$. We use the splits provided by \cite{tran2024ufsa}. A comparison to the state of the art using this protocol for the Breakfast \cite{kuehne2014breakfast} and YouTube Instructional \cite{alayrac2016narrated} datasets is provided in \tabref{tab:breakfast_results_supp} and \tabref{tab:yti_results_supp}, respectively. Our approach outperforms the state of the art also for this protocol on both datasets.       

\begin{table}[tb]
        \caption{Results on Breakfast using 80:20 split. Best results are in \textbf{bold}, while second best ones are \underline{\textit{underlined}}.}
    
        \centering
        \footnotesize
        \begin{tabular}{c@{\hspace{1em}}cc}
            \toprule
            \textbf{Method}  &\textbf{MOF}  &\textbf{F1}\\
            \midrule
                     
            CTE~\cite{kukleva2019unsupervised} & 39.8 & 25.5\\            
            TOT~\cite{kumar2022tot} & 40.6 & 27.6 \\
            TOT+TCL~\cite{kumar2022tot} & 37.4 & 23.2 \\
            UFSA~\cite{tran2024ufsa} & \underline{\it 44.0} & \underline{\it 36.7} \\
            Ours (\ours) & {\bf 44.2} & {\bf 37.0} \\
            
            \hline        
            \label{tab:breakfast_results_supp}    
        \end{tabular}
\end{table}

\begin{table}[tb]
        \caption{Results on YouTube Instructions using 80:20 split. Best results are in \textbf{bold}, while second best ones are \underline{\textit{underlined}}.}
        
        \centering
        \footnotesize
        \begin{tabular}{c@{\hspace{1em}}cc}
            \toprule
            \textbf{Method} &\textbf{MOF}  &\textbf{F1}\\
            \bottomrule
             
            CTE~\cite{kukleva2019unsupervised} & 38.4 & 25.5\\
            TOT~\cite{kumar2022tot} & 40.4 & 28.0 \\
            TOT+TCL~\cite{kumar2022tot} & 40.6 & 26.7 \\
            UFSA~\cite{tran2024ufsa} & \underline{\it 46.8} & \underline{\it 28.2} \\
            Ours (\ours) & {\bf 50.7} & {\bf 39.6} \\
            
            \hline        
            \label{tab:yti_results_supp}
        \end{tabular}   

\end{table}

\section{Qualitative Results}
\label{sec:qualitative}

\begin{figure}[tb]
    \centering
    \includegraphics[width=1\linewidth]{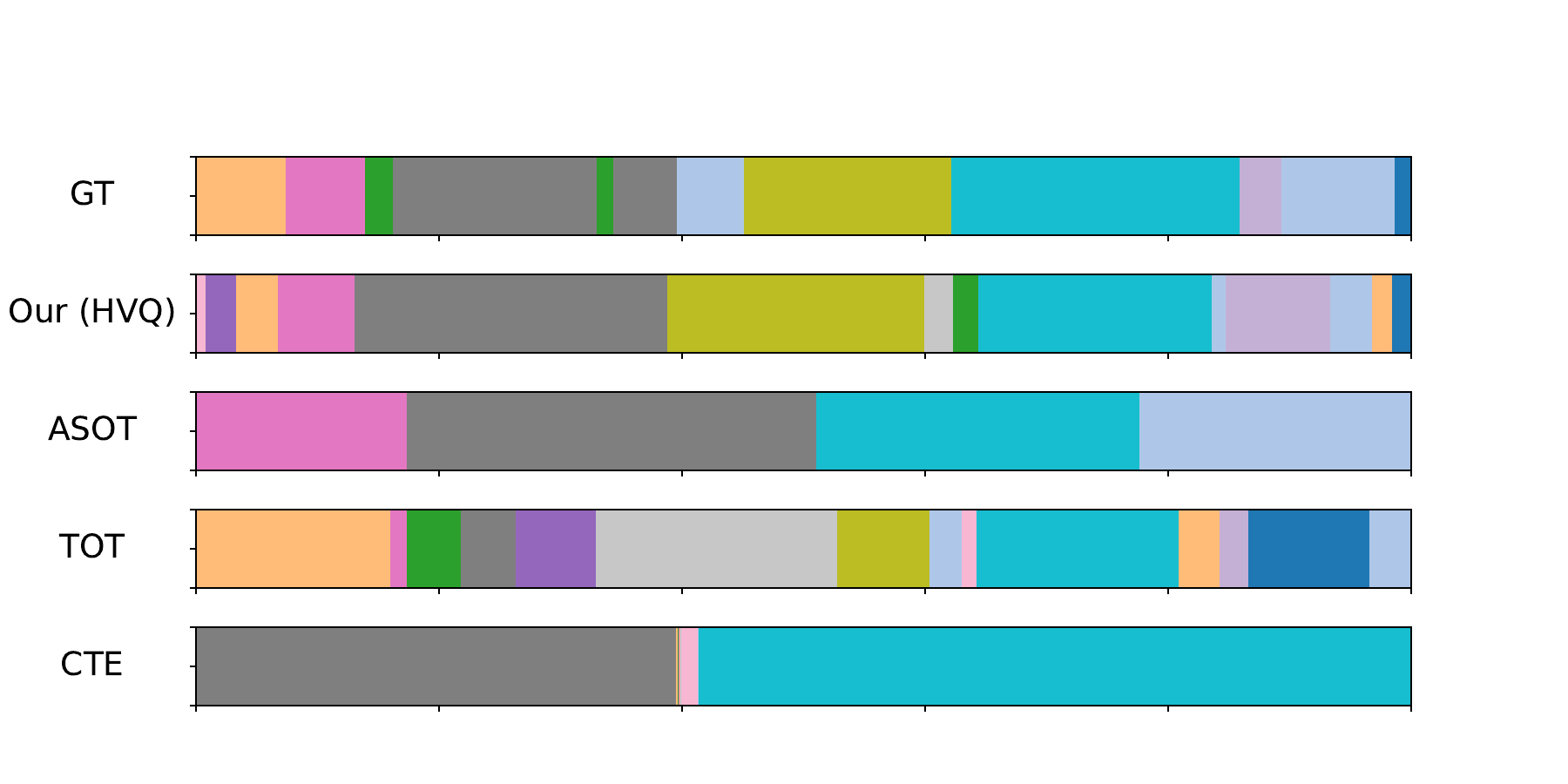}
    \caption{Segmentation results for video \textit{P23\_webcam01\_pancake} of Breakfast.}
    \label{fig:pancake}
\end{figure}

\figref{fig:pancake} shows qualitative results for a video belonging to the activity {\it pancake} of Breakfast. The activity {\it pancake} contains short and long action segments. 
Our model shows its capability to handle both short and long actions. ASOT predicts only long segments. TOT does not recognize the long dark gray action. Our method is the only one that successfully segments the last blue segment, the initial pink one and the middle yellow-green one.   

\begin{figure}[tb]
    \centering
    \includegraphics[width=1\linewidth]{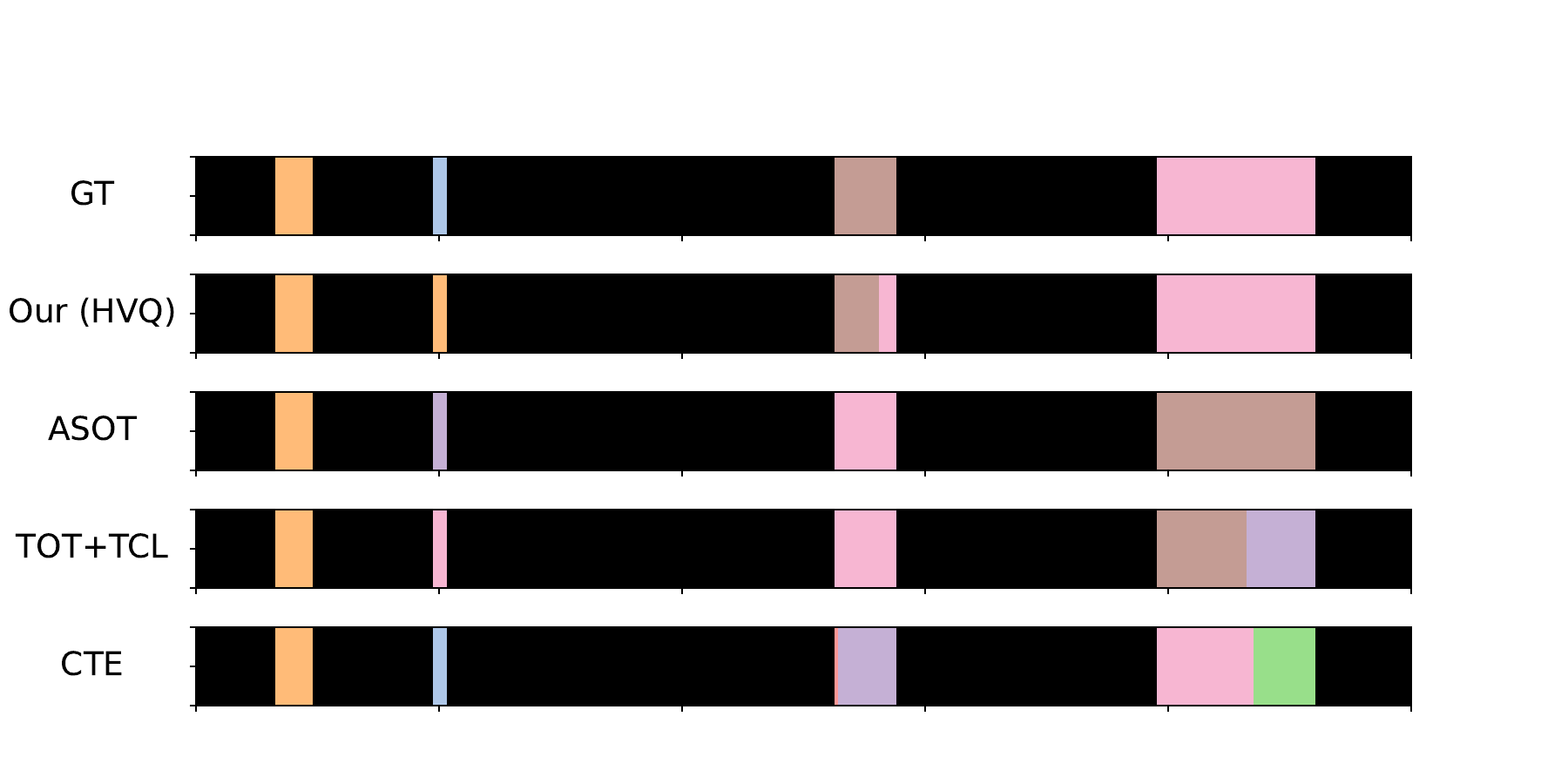}
    \caption{Segmentation results for video {\it cpr\_0023} of YouTube Instructional. Black color indicates background frames.}
    \label{fig:cpr}
\end{figure}

\begin{figure}[tb]
    \centering
    \includegraphics[width=1\linewidth]{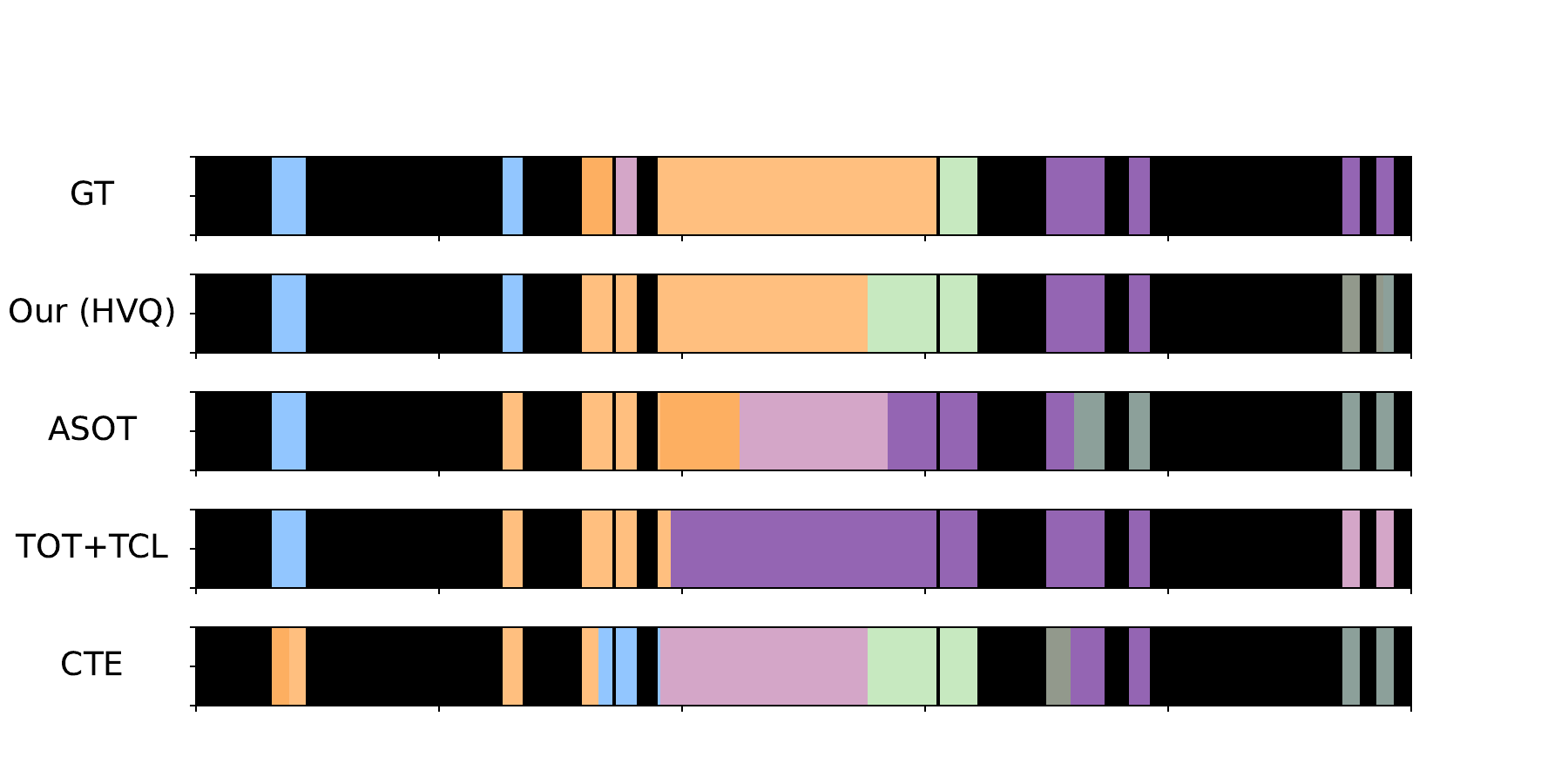}
    \caption{Segmentation results for video {\it repot\_0003} of YouTube Instructional. Black color indicates background frames.}
    \label{fig:repot}
\end{figure}

We represent qualitative results for the YouTube Instructional dataset in the same way as \cite{kumar2022tot, tran2024ufsa}, by superposing the background mask on the predictions, in Figs.~\ref{fig:cpr}-\ref{fig:repot}. In \figref{fig:cpr}, our method is the only approach that correctly segments and classifies the last action, {\it give compression}, and almost correctly identifies the dark-orange action ({\it give breath}). 
In \figref{fig:repot}, our method is the only approach that is capable to correctly label the longest light-orange action in the middle, {\it loosen root}, and the initial blue action {\it put soil}.

\begin{figure}[tb]
    \centering
    \includegraphics[width=1\linewidth]{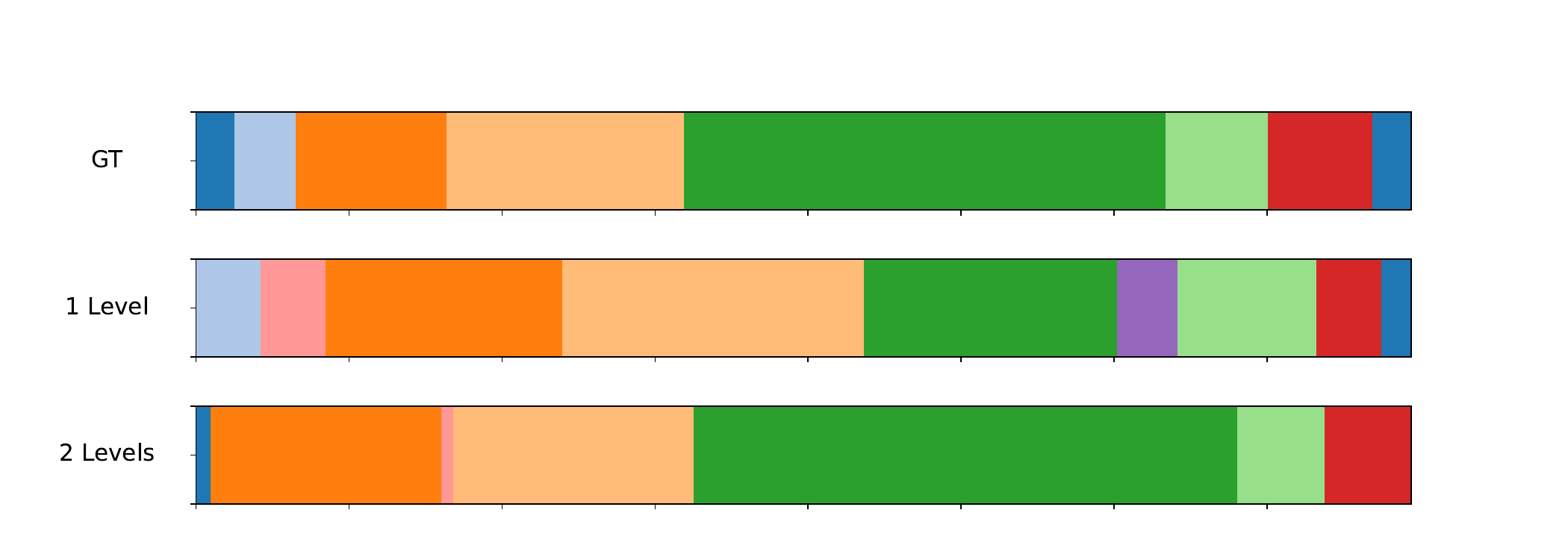}
    \caption{Segmentation results for video {\it P22\_cam01\_friedegg} of Breakfast. `1 Level' refers to our approach with one level, while `2 Levels' refers to our model with two levels.}  
    \label{fig:hie-fried1}
\end{figure}

\begin{figure}[tb]
    \centering
    \includegraphics[width=1\linewidth]{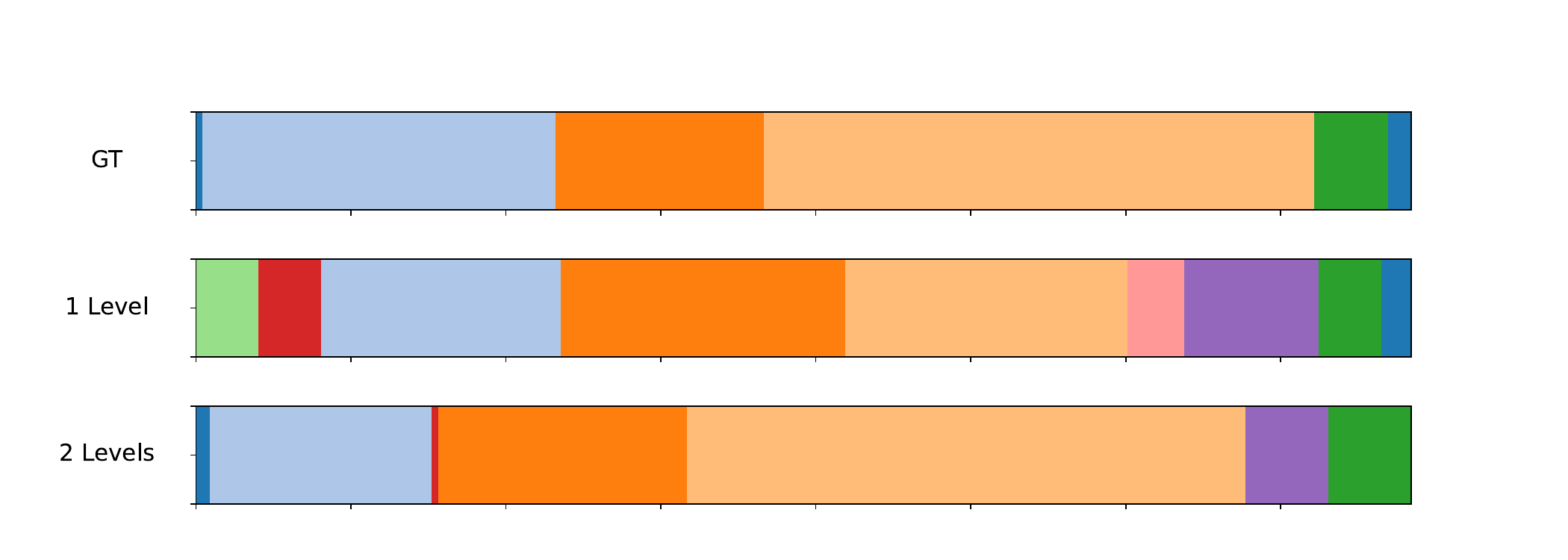}
    \caption{Segmentation results for video {\it P45\_webcam01\_friedegg} of Breakfast. `1 Level' refers to our approach with one level, while `2 Levels' refers to our model with two levels.}
    \label{fig:hie-fried2}
\end{figure}

\begin{figure}[tb]
    \centering
    \includegraphics[width=1\linewidth]{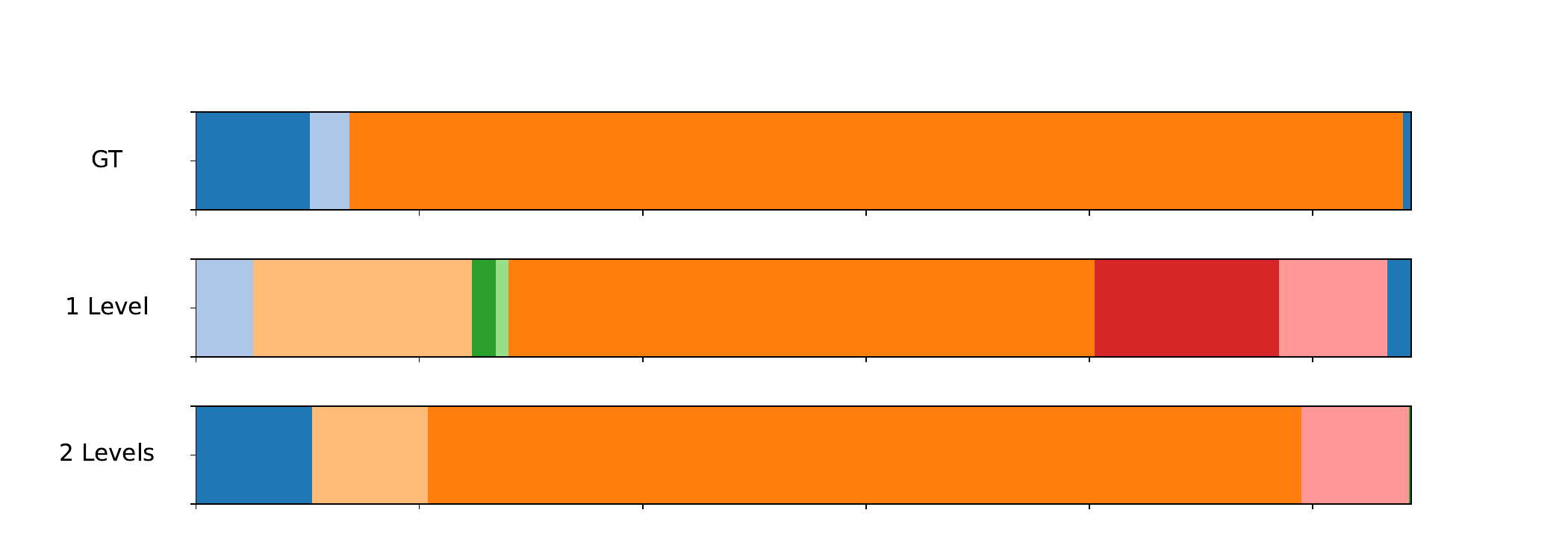}
    \caption{Segmentation results for video {\it P45\_stereo01\_salat} of Breakfast. `1 Level' refers to our approach with one level, while `2 Levels' refers to our model with two levels.}
    \label{fig:hie-salat}
\end{figure}

Figs.~\ref{fig:hie-fried1}, \ref{fig:hie-fried2}, and \ref{fig:hie-salat} show qualitative comparisons between hierarchical vector quantization and vector quantization with one level on Breakfast. The examples show that hierarchical vector quantization is able to model longer actions better (\eg, the dark-orange action, {\it cut fruit}, in \figref{fig:hie-salat}).

\end{document}